\documentclass[conference]{IEEEtran}
\IEEEoverridecommandlockouts

\usepackage{cite}
\usepackage{amsmath,amssymb,amsfonts}
\usepackage{algorithmic}
\usepackage{graphicx}
\usepackage{textcomp}
\usepackage{xcolor}
\usepackage{placeins}
\usepackage{stfloats}
\usepackage{booktabs}
\usepackage{float}

\usepackage{enumitem} % no preamble

\usepackage{array}
\usepackage{amssymb}
\def\BibTeX{{\rm B\kern-.05em{\sc i\kern-.025em b}\kern-.08em
    T\kern-.1667em\lower.7ex\hbox{E}\kern-.125emX}}

\begin{document}

\bstctlcite{IEEEexample:BSTcontrol} 

\title{Toward a Decision-Assurance Layer for AI-Assisted Flight Planning in Air Traffic Management}

\author{\IEEEauthorblockN{1\textsuperscript{st} Alexandre Barreto}
\IEEEauthorblockA{\textit{Department of Cyber Security Engineering} \\
\textit{George Mason University}\\
Fairfax, VA, USA \\
adebarro@gmu.edu}
\and
\IEEEauthorblockN{2\textsuperscript{nd} Shou Matsumoto}
\IEEEauthorblockA{\textit{C5I Center} \\
\textit{George Mason University}\\
Fairfax, VA, USA \\
smatsum2@gmu.edu}
\and
\IEEEauthorblockN{3\textsuperscript{rd} Jorge Valverde-Rebaza}
\IEEEauthorblockA{\textit{Department of Computer Science} \\
\textit{Tecnológico de Monterrey}\\
Mexico City, Mexico \\
jvalverr@tec.mx}
\and
\IEEEauthorblockN{4\textsuperscript{th} Cleiton Ataide}
\IEEEauthorblockA{\textit{Technical Subdepartment} \\
\textit{Department of Airspace Control (DECEA)}\\
Rio de Janeiro, Brazil \\
ataidecaa@decea.mil.br}
\and
\IEEEauthorblockN{5\textsuperscript{th} Paulo Costa}
\IEEEauthorblockA{\textit{C5I Center} \\
\textit{George Mason University}\\
Fairfax, VA, USA \\
pcosta@gmu.edu}
}

\maketitle

\begin{abstract}
Generative AI is increasingly being used informally in Air Traffic Management (ATM) for tasks such as flight plan generation, trajectory interpretation, and constraint checking. 
\mbox{Although} these tools can reduce workload and accelerate planning, their non-deterministic outputs create safety and operational risks in human-in-the-loop settings. 
This paper proposes the AI Trust and Assurance Layer (ATAL), a model-agnostic decision assurance architecture that evaluates whether AI-generated flight-planning outputs are sufficiently reliable for operational use. 
ATAL combines semantic stability under prompt variation, operational consistency of structured outputs, and normative constraint validation against domain rules, and maps these signals to a Decision Readiness Level (DRL) for human operators. 
An ATM-inspired experimental study shows how unsafe, inconsistent, or misleading outputs can be identified before influencing flight-plan validation or execution.
Although demonstrated in aviation, the framework is also transferable to other safety-critical decision-support domains that require human oversight under regulatory constraints.
\end{abstract}


\section{Introduction}

We are witnessing a rapid transition of Frontier AI from conversational chatbots to autonomous decision-support agents \cite{qin2025, yao2023, park2023}. While Large Language Models (LLMs) demonstrate remarkable reasoning capabilities, their reliance on statistical patterns remains a fundamental hurdle for safety-critical applications \cite{brown2020, rafael2020}. In domains where precision is non-negotiable, the inherent unpredictability of probabilistic systems is not merely a technical flaw—it is a barrier to operational trust.

LLMs show strong reasoning skills, but some built-in traits make safe use difficult. Their outputs can change unpredictably, they react to small changes in input, they may seem overly confident without showing uncertainty, and they are open to adversarial attacks \cite{brown2020, zou2023, liu2025,Damacena2026}. These features are natural to probabilistic generative systems, not flaws \cite{brown1992,ji2023}. Still, when used in decision-making, these traits can lead to a lack of trust.

Current evaluation methods mostly use fixed benchmarks and human preferences, but these are not enough when reliability is critical, such as during distribution changes, adversarial attacks, or in complex situations \cite{srivastava2023, liang2023}. Traditional methods do not provide the needed confidence for decisions or predictions, making it harder to oversee models that go beyond standard testing \cite{lin2022, perez2022}.

Despite these limitations, the potential for LLMs to augment human expertise in complex environments like aviation is undeniable. Flight planning, for instance, demands a rigorous synthesis of regulatory constraints, specialized codes, and real-time data—a process in which even seasoned pilots can experience cognitive overload as rules evolve. In the ATM context, a Flight Plan (FPL) is not merely a statement of routing intent; it is a binding operational contract. It synthesizes aircraft performance, airspace restrictions, and adherence to International Civil Aviation provisions (ICAO) \cite{ICAO44442016}. GenAI-driven automation must ensure that FPL syntax—specifically Items 10, 15, and 18—precisely reflects real-world equipment and performance limitations to maintain safety.

This creates a central challenge for ATM: AI can accelerate flight planning, but its probabilistic and opaque behavior is difficult to audit using traditional safety methods \cite{rtca2011, nistai2023, euai2024, iec2010}. 
To address this gap, we propose the \textbf{AI Trust and Assurance Layer (ATAL)}, a model-agnostic framework that evaluates whether AI-generated outputs are stable, operationally consistent, and normatively compliant before they can inform execution decisions.

ATAL acts as an assurance layer between LLM-generated advisories and operational use. It combines semantic stability, operational consistency, and normative constraint validation, and maps these signals to a \textbf{Decision Readiness Level (DRL)} that supports calibrated human oversight. 
We demonstrate the framework in an ATM case study on flight plan generation and validation. Section~\ref{sec:Literature Review} reviews related work, Section~\ref{sec:AI Trust and Assurance Layer (ATAL)} presents ATAL and DRL, Section~\ref{sec:Experimental Results and Analysis} reports the experimental evaluation, and Section~\ref{sec:Conclusion} concludes the paper.

\section{Literature Review}
\label{sec:Literature Review}

Traditional evaluation of language models has relied on probabilistic metrics such as perplexity \cite{brown1992}. While useful for gauging linguistic fluency, perplexity is a poor proxy for safety. A model can assign high probability to a syntactically flawless yet operationally catastrophic sequence. In safety-critical domains, 'fluency' is secondary to structural validity and regulatory compliance—areas where token-level uncertainty metrics provide little to no visibility.

Modern frameworks like HELM and TruthfulQA have indeed advanced our ability to quantify hallucinations and factual consistency \cite{ji2023}. However, these benchmarks remain largely rooted in general-knowledge retrieval. They fail to account for the deterministic rigors of specialized domains. Knowing a fact is fundamentally different from adhering to a set of interlocking aviation constraints where distribution shifts or adversarial prompts can trigger unpredictable failures.

It is important to note that none of these metrics address whether a model output is sufficiently reliable to support time-critical decisions under regulatory limitations.

Current safety paradigms rely heavily on behavioral alignment and benchmark performance, but these remain insufficient for safety-critical deployment \cite{christiano2017, Ouyang2022}. 
A model may appear helpful and well aligned while still exhibiting structural unreliability, overconfidence \cite{guo2017}, brittleness under distribution shift \cite{hendrycks2017}, or vulnerability to adversarial prompting \cite{goodfellow2015}. 
As a result, existing methods do not provide a predictive basis for determining whether apparently safe behavior in controlled settings will remain reliable in real operational loops. 
This gap motivates the need for an assurance framework that evaluates behavioral stability, constraint compliance, and decision readiness together rather than in isolation \cite{nistai2023}.

In parallel, traditional safety engineering uses assurance cases and formal validation techniques to demonstrate compliance with safety standards (e.g., DO-178C in aviation \cite{rtca2011}). These methods rely on deterministic models and traceable requirements.

Recent work in AI assurance has explored explainability methods, statistical uncertainty assessment, and governance-level risk mitigation frameworks \cite{liang2023, ji2023, nistai2023}. 

However, these approaches either focus on model explainability and statistical confidence or assume deterministic system operation. They do not integrate structured domain constraints, perturbation-based semantic stability testing, adversarial diagnostics, and calibration assessment into an integrated operational metric that supports graded decision authority.

\FloatBarrier
\begin{table*}[t]
\centering
\caption{Limitations of Existing AI Evaluation Approaches and ATAL Contributions}
\label{tab:atal_gap}

\scriptsize
\setlength{\tabcolsep}{4pt}
\renewcommand{\arraystretch}{1.2}

\begin{tabular}{p{2.2cm} p{3cm} p{4.2cm} p{3.5cm} p{3.5cm}}
\hline
\textbf{Category} & 
\textbf{Capability} & 
\textbf{Limitation (Gap)} & 
\textbf{Existing Approaches} & 
\textbf{ATAL Contribution} \\
\hline

Behavioral Evaluation 
& Linguistic quality and fluency 
& Focuses on surface-level correctness without ensuring semantic or structural reliability 
& Perplexity, GLUE, HELM 
& Not primary focus; used as auxiliary signal \\

\hline

Factual Consistency 
& Hallucination detection 
& Detects factual errors but lacks structural validation and cross-field consistency 
& TruthfulQA, factuality benchmarks 
& Partial detection via behavioral inconsistency analysis \\

\hline

Constraint Enforcement 
& Domain and rule validation 
& Limited to syntactic or schema-level validation; lacks semantic and contextual enforcement 
& Guardrails, schema validation 
& Deterministic and contextual constraint validation (NCV) \\

\hline

Robustness 
& Stability under perturbations 
& No systematic evaluation of output invariance under controlled input variations 
& Rarely addressed explicitly 
& Semantic Stability Score (SSS) for perturbation-based robustness \\

\hline

Adversarial Resilience 
& Detection of adversarial manipulation 
& Isolated detection without integration into reliability assessment 
& Adversarial prompt filters 
& Integrated adversarial signal detection (IEI) \\

\hline

Operational Monitoring 
& Longitudinal degradation tracking 
& Monitoring exists but disconnected from decision-making and reliability scoring 
& Drift detection systems 
& Structured longitudinal monitoring (TRL layer) \\

\hline

Decision Support 
& Decision-readiness assessment 
& No unified metric to assess operational reliability of AI outputs 
& Conceptual frameworks (e.g., NIST AI RMF) 
& Decision Readiness Level (DRL 1--4) aggregation \\

\hline

System Integration 
& End-to-end evaluation 
& Fragmented tools without unified framework or interoperability 
& All existing approaches 
& Unified black-box compatible evaluation layer \\

\hline
\end{tabular}
\end{table*}

Across these research streams, a common limitation remains: existing methods address individual aspects of reliability but do not provide a unified, domain-aware layer that evaluates semantic invariance, structural consistency, deterministic constraints, adversarial signals, and longitudinal degradation in a single interpretable measure of decision readiness. ATAL is proposed to fill this gap (see Table~\ref{tab:atal_gap}).

\section{AI Trust and Assurance Layer (ATAL)}
\label{sec:AI Trust and Assurance Layer (ATAL)}

\textbf{AI Trust and Assurance Layer (ATAL)} is a black-box, domain-aware assurance framework that measures whether frontier AI outputs remain semantically stable, operationally consistent, and normatively valid under realistic perturbations before they are used in high-stakes workflows.

We implement ATAL as a multi-dimensional measurement suite that serves as a rigorous 'behavioral filter' for frontier AI agents. Our architecture does not merely observe; it subjects model outputs to a battery of stress tests across five analytical layers to synthesize a high-integrity oversight signal. The constituent elements of the ATAL framework include:

\begin{itemize}[leftmargin=*, itemsep=6pt]

\item \textbf{Five analytical dimensions:}
\begin{enumerate}[leftmargin=*, itemsep=4pt, parsep=2pt]

\item \textbf{[Core] Semantic Stability (SSS):} A leading indicator that evaluates the invariance of underlying semantics across diverse prompt formulations. Its role is to detect meaning drift under controlled paraphrasing and perturbations.

\item \textbf{[Core] Operational Consistency (OCS):} Assesses whether structured operational decisions remain invariant across outputs. In conjunction with SSS, it captures decision-level drift under semantic and parametric variations.

\item \textbf{[Core] Normative Constraint Validation (NCV):} Acts as a ground-truth mediator by enforcing compliance with deterministic domain constraints (e.g., FAA separation minima), enabling correlation between semantic instability and rule violations.

\item \textbf{[Ancillary] Injection Evidence Index (IEI):} A robustness layer that detects adversarial manipulation signals, including prompt injection and jailbreak attempts, and evaluates their impact on semantic and operational consistency.

\item \textbf{[Ancillary] Temporal Reliability Layer (TRL):} A longitudinal monitoring component that tracks systemic degradation over time, identifying reliability drift under repeated usage or shifting data distributions.

\end{enumerate}

\item \textbf{One synthesized composite metric:}
\begin{enumerate}[leftmargin=*, itemsep=4pt, parsep=2pt, start=1]

\item \textbf{[Output] Decision Readiness Level (DRL):} We propose this metric as an empirical threshold and gatekeeping mechanism governing the transition from model output to actionable decisions for human or autonomous systems. This mechanism is intended to support calibrated oversight in high-stakes environments. For operational clarity, the composite output is mapped into a four-tier classification schema: \textbf{DRL-1} (unsuitable for use), \textbf{DRL-2} (informational/reference/read-only), \textbf{DRL-3} (conditional execution subject to human supervision), and \textbf{DRL-4} (fully actionable/decision-ready). 
This taxonomy provides a streamlined structure to support rapid decision-making in critical settings.

\end{enumerate}

\end{itemize}

\begin{figure*}[t]
\centering
\includegraphics[width=.98\textwidth]{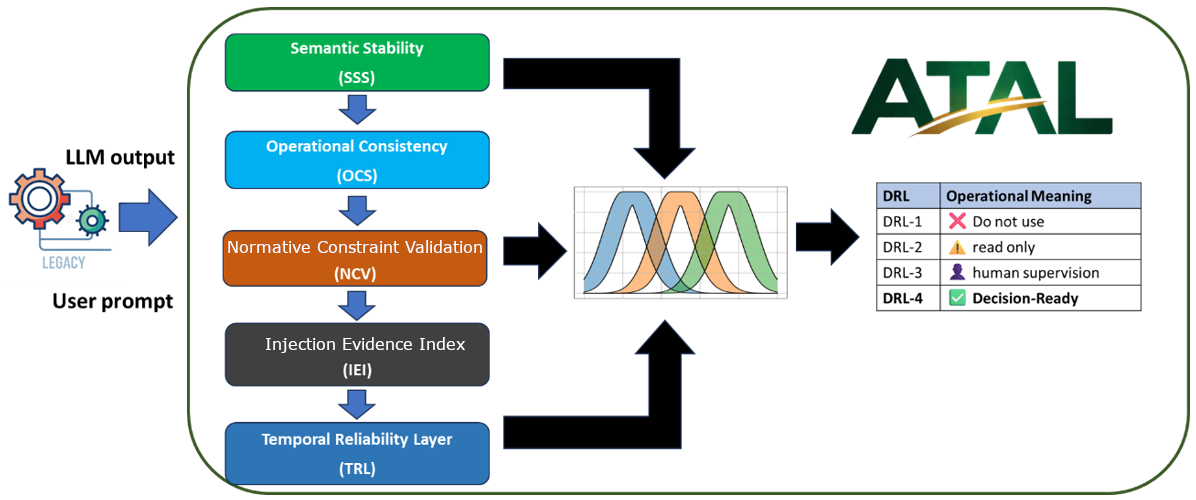}
\caption{The ATAL Framework Architecture.}
\label{fig:atal_architecture}
\end{figure*}

Outputs from the DRL framework will extend beyond simple classification labels, providing additional high-utility insights for stakeholders. Specifically, each output will be augmented with quantified uncertainty metrics, such as fuzzy-logic membership degrees or full probability distributions, alongside interpretability/explainability features designed to characterize confidence and facilitate trust \cite{arrieta2020, dwivedi2023, gunning2021}. This comprehensive data packet is engineered to support both nuanced human-in-the-loop oversight and the seamless execution of autonomous protocols.

While ATAL may provide robust evaluative capabilities, it is subject to specific boundary conditions. The framework is not designed to certify absolute factual correctness or to detect deeply hidden misalignments that do not manifest behaviorally. Its utility is optimized for domains in which formal constraints are explicitly defined; therefore, safety may not generalize to ungrounded environments. Additionally, the system's resilience is bounded by the validated constraint models and does not guarantee extensibility to novel attack vectors or out-of-distribution operational scenarios.

The ATAL architecture is designed inherently to support resilience against adversarial prompt injection (IEI) and longitudinal quality degradation (TRL) through its modular evaluation layers. 
However, the core instantiated components in this study are SSS, OCS, NCV, and the DRL aggregation module. The IEI and TRL are included in the broader architecture but remain only partially implemented and are left for future evaluation.

\subsection{Semantic Stability (SSS)}
\label{SSS}

The Semantic Stability Score (SSS) quantifies the robustness of a language model's output to controlled, meaning-preserving prompt variations. Its core objective is to assess whether small, semantically equivalent reformulations of an input prompt lead to consistent operational outcomes. 

\begin{figure}[htbp]
\centering
    \includegraphics[width=1.02\columnwidth]{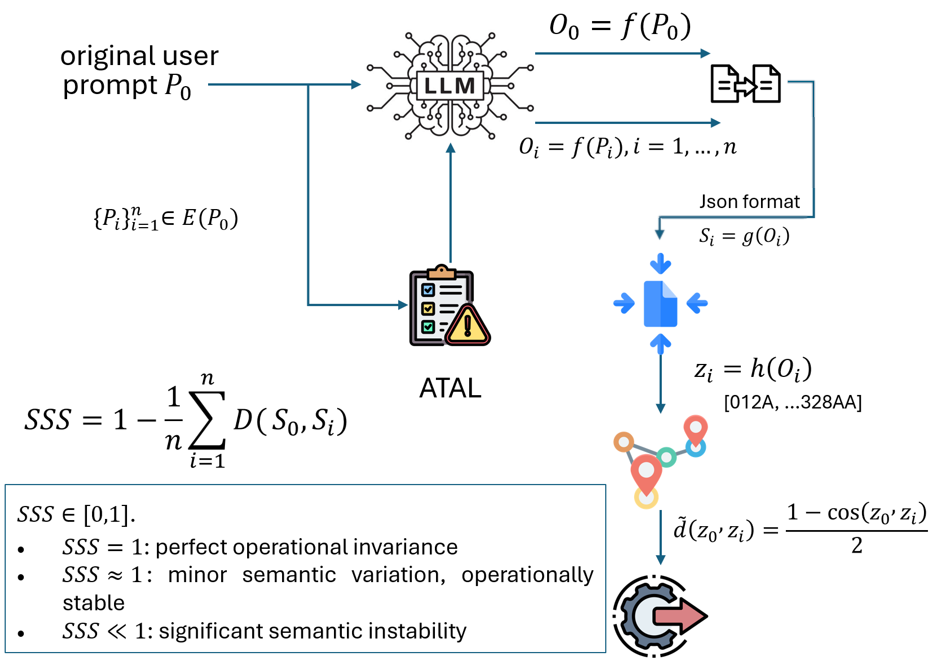}
    \caption{Semantic Stability Score (SSS) Pipeline}
    \label{fig:sss_pipeline}
\end{figure}

In this sense, SSS captures semantic invariance, rather than linguistic similarity, focusing on whether the decision encoded in the output remains stable despite superficial changes in wording.
Figure \ref{fig:sss_pipeline} illustrates how SSS evaluates semantic stability by generating controlled, intent-preserving prompt variations and comparing the resulting LLM outputs to assess operational invariance, raising a semantic drift flag when meaning or structured parameters diverge beyond a defined threshold.

Given an original prompt $P_0$ and a set of semantically equivalent variations $\{P_i\}_{i=1}^{n}$, the model produces corresponding outputs $o_i$. Each output is then transformed into a structured representation $S_i = g(o_i)$, where $g(\cdot)$ maps raw text into a domain-specific semantic schema. The Semantic Stability Score is defined as the complement of the average divergence between the reference output $S_0$ and each perturbed output $S_i$, i.e., it measures how much the model's behavior deviates across equivalent inputs.

The divergence function $D(S_0, S_i)$ captures differences in the semantic representation of outputs. In practice, this is implemented using continuous similarity measures, such as embedding-based or structured similarity measures over the semantic representation. By operating on structured representations rather than raw text, SSS ensures that evaluation focuses on \textbf{operational meaning} rather than surface-level phrasal variations. As a result, two outputs that are linguistically different but operationally equivalent will exhibit low divergence, while small semantic shifts that affect decisions will be penalized.

The resulting SSS value lies in the interval $[0,1]$ where values close to 1 indicate high semantic stability, i.e., the model consistently preserves the same operational intent across prompt reformulations, whereas lower values indicate increasing levels of semantic fragility. A low SSS suggests that the model is sensitive to minor input perturbations, potentially leading to inconsistent or unreliable decisions in safety-critical contexts.

Importantly, SSS is not a measure of correctness or compliance with domain rules. Instead, it isolates the notion of \textbf{consistency under perturbation}, enabling a causal separation between input variability and model behavior. This makes SSS a foundational component in broader assurance frameworks, where it is combined with complementary metrics, such as structural consistency and normative validation, to assess overall decision readiness.

To ensure the quality of perturbations derived from the user prompt, we use the \textbf{Variation Quality Score (VQS)}. This score has two main objectives: \textit{(i)} to verify that generated prompt variations preserve the original semantic intent and critical domain entities, and \textit{(ii)} to filter out malformed, irrelevant, or structurally inconsistent perturbations that could invalidate the evaluation.

VQS is defined as the ratio between the number of valid variations and the total number of generated variations. A variation is considered valid if it satisfies semantic-similarity constraints relative to the original prompt and preserves key anchors (e.g., domain-specific entities such as locations, identifiers, or parameters). By acting as a gating mechanism, VQS ensures that only high-quality, intent-preserving perturbations are used to compute the Semantic Stability Score (SSS).

As a result, VQS enables a clear causal separation between input quality and model behavior: low VQS indicates an unreliable perturbation process, while low SSS (under high VQS) indicates genuine semantic instability of the model.

The joint interpretation of VQS and SSS provides deeper insight into model performance. For instance, when both VQS and SSS are high, the model can be considered robust and reliable, as it produces consistent outputs under high-quality perturbations. However, when VQS is high, but SSS is low, the model exhibits genuine semantic instability, indicating sensitivity to minor input changes. 
Table~\ref{tab:vqs_sss} summarizes the qualitative interpretations of VQS and SSS.

\begin{table}[ht]
\centering
\caption{Joint Interpretation of VQS and SSS}
\label{tab:vqs_sss}
\begin{tabular}{c c l}
\toprule
\textbf{VQS} & \textbf{SSS} & \textbf{Interpretation} \\
\midrule
High & High & Model is robust and reliable \\
High & Low  & Model is unstable under valid variations \\
Low  & Low  & Invalid experiment (bad perturbations) \\
Low  & High & Model is robust even under noisy input \\
\bottomrule
\end{tabular}
\end{table}

Overall, the combined use of VQS and SSS establishes a principled evaluation framework that avoids misleading conclusions and strengthens the validity of robustness assessments. By explicitly decoupling perturbation quality from model response consistency, this approach provides a more reliable foundation for analyzing the behavior of language models in safety-critical applications.

\subsection{Operational Consistency Score (OCS)}
\label{OCS}

While SSS looks at the input-output relationship, the \textbf{Operational Consistency Score (OCS)} looks at the 'internal consensus' of the system. We designed OCS to detect structural contradictions between multiple advisories. By mapping raw text to domain-specific attributes, we can mathematically verify if two differently worded flight plans actually propose the same trajectory, ensuring that 'fluency' never masks a fundamental change in operational intent.

The core question addressed by OCS is: do multiple advisory outputs represent the same operational decision when mapped into a structured representation? This formulation enables the comparison of AI outputs at the level of operational intent rather than surface-level textual similarity, ensuring that semantically equivalent decisions are recognized as consistent even when expressed differently. Operational consistency is validated by extracting key attributes from the generated plan, such as Field 15 (Route) and Field 18 (e.g., PBN/, NAV/) \cite{ICAO44442016}. The OCS ensures that despite natural language variations in the prompt, the resulting trajectory and capability codes remain invariant. This prevents equipment code 'hallucinations' that would lead to the rejection of the plan by ground-based processing systems.

\begin{figure}[htbp]
    \centering
    \includegraphics[width=.99\columnwidth]{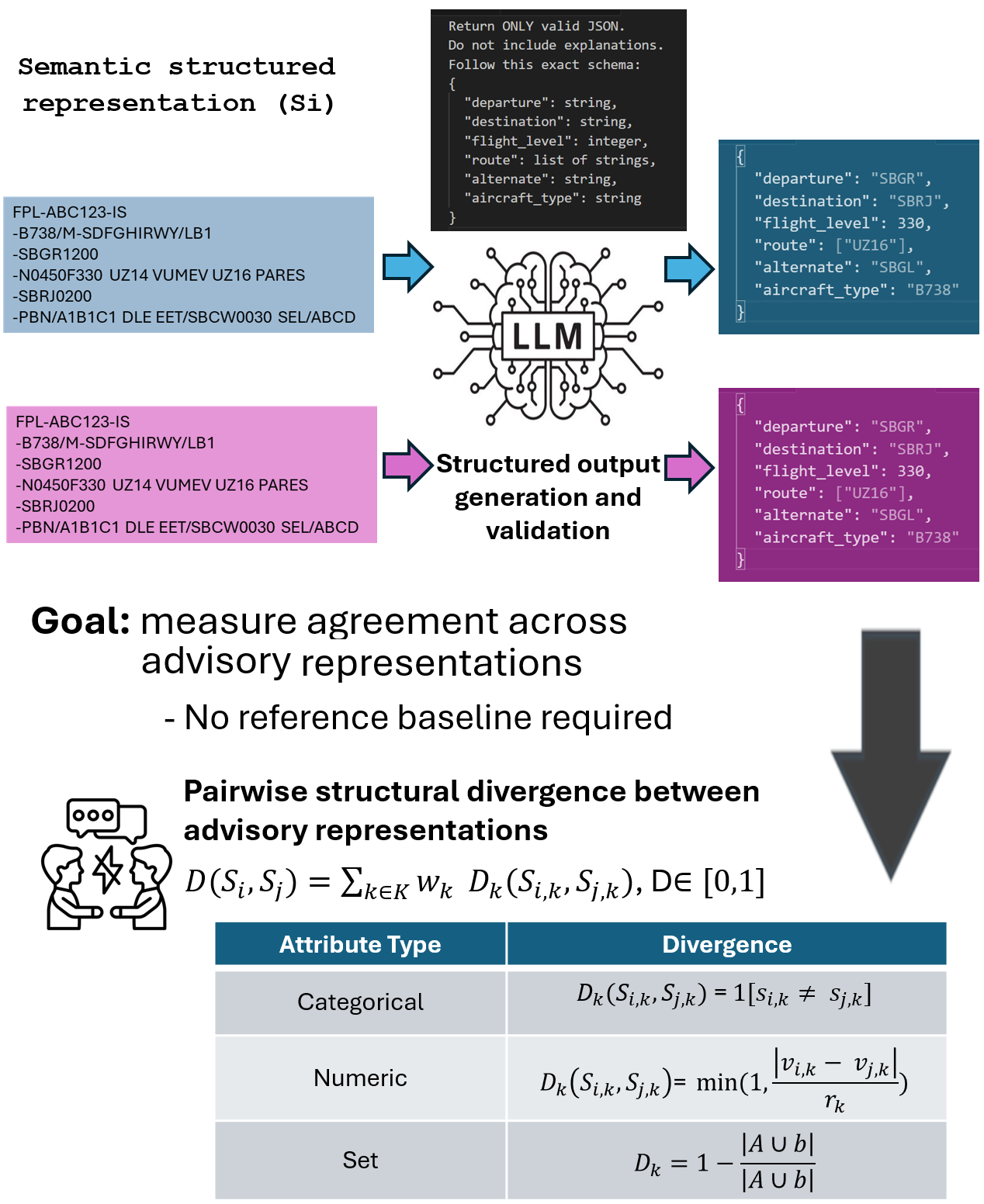}
    \caption{Operational Consistency Score (OCS) pipeline.}
    \label{fig:ocs_pipeline}
\end{figure}

Figure~\ref{fig:ocs_pipeline} illustrates how advisory outputs are transformed into structured representations and compared through attribute-level divergence to compute pairwise consistency.

Let a set of $n$ advisory outputs be given, each transformed into a structured representation $S_i$, where $S_i$ consists of a set of domain-specific attributes (e.g., departure, destination, flight level, route, and aircraft type). The pairwise structural divergence between two advisories $S_i$ and $S_j$ is defined as:
\[
D(S_i, S_j) = \sum_{k \in K} w_k \, D_k(s_{i,k}, s_{j,k}),
\] 
where $K$ is the set of monitored attributes, $w_k$ are normalized weights such that $\sum_{k \in K} w_k = 1$.

The divergence function $D_k$ depends on the attribute type. For categorical attributes, $D_k = \mathbb{1}[s_{i,k} \neq s_{j,k}]$. For numerical attributes, $D_k = \min\left(1, \frac{|v_{i,k} - v_{j,k}|}{r_k}\right)$, where $r_k$ is a normalization factor. For set-valued attributes, $D_k = 1 - \frac{|A \cap B|}{|A \cup B|}$.

Let $M = \frac{n(n-1)}{2}$ denote the number of unordered pairs. The Operational Consistency Score is defined as:
\[
OCS = 1 - \frac{1}{M} \sum_{i<j} D(S_i, S_j), \quad OCS \in [0,1].
\]

An OCS value of 1 indicates perfect structural agreement across advisories, meaning that all outputs encode identical operational parameters. Values close to 1 indicate minor variations that do not affect the underlying decision, whereas lower values indicate increasing divergence across advisories. When OCS falls below a predefined threshold, an operational inconsistency condition may be triggered.

OCS supports two evaluation contexts: (i) multiple AI advisors providing recommendations for the same scenario, and (ii) multiple executions of a single advisor under identical or similar conditions. In both cases, OCS quantifies agreement across outputs independently of correctness, thereby complementing semantic stability and normative validation within the ATAL framework.

\subsection{Normative Constraint Validation (NCV)}
\label{NCV}

The \textbf{Normative Constraint Validation (NCV)} layer serves as our ground-truth anchor. In aviation, safety is not subjective. We built NCV to enforce a hierarchical gatekeeping process: \textbf{Hard Safety Constraints} (like separation minima) act as binary switches—any violation immediately invalidates the output. \textbf{Contextual and soft constraints} then refine this score, penalizing inefficiency without compromising the 'safe-to-fail' nature of the system.

Unlike Semantic Stability (SSS), which measures behavioral sensitivity under input perturbations, and Operational Consistency Score (OCS), which assesses structural agreement across outputs, NCV evaluates objective correctness with respect to domain-specific normative rules.

Given a structured advisory representation \( S \), constraints are organized into three layers:

\begin{enumerate}
    \item \textit{Hard Safety Constraints (HSC)} — mandatory rules whose violation immediately invalidates the advisory, including cross-referencing against current Route Availability Document (RAD) restrictions. For instance, an AI-generated plan that ignores a No-Fly Zone or utilizes an airway in the incorrect direction of flight is immediately invalidated, regardless of its linguistic fluency or perceived confidence; 
    \item \textit{Contextual Constraints (CC)} — conditions dependent on external or situational factors; and 
    \item \textit{Soft Constraints (SC)} — non-critical conditions reflecting operational quality or efficiency.
\end{enumerate}

\begin{figure}[!htbp]
    \centering
    \includegraphics[width=1.02\columnwidth]{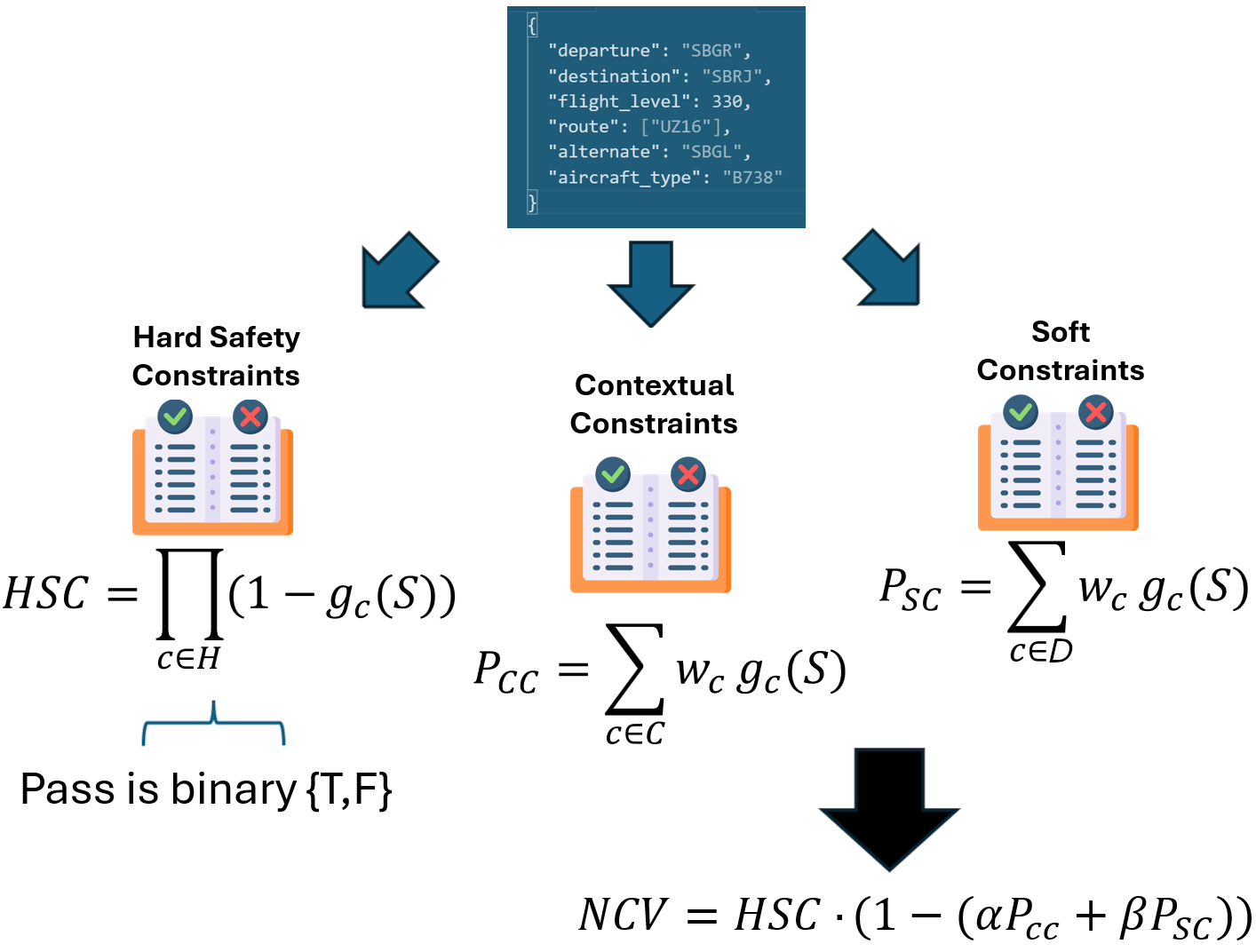}
    \caption{Normative Constraint Validation (NCV).}
    \label{fig:ncv_pipeline}
\end{figure}

Figure~\ref{fig:ncv_pipeline} illustrates the layered evaluation process, where hard safety constraints act as a binary feasibility gate, and contextual and soft constraints introduce weighted penalties.

Let \( g_c(S) \in \{0,1\} \) indicate whether constraint \( c \) is violated. For a set of hard constraints $H$, the hard constraint compliance indicator is defined as:
\[
HSC = \prod_{c \in H} (1 - g_c(S))
\]

For a set of contextual constraints $\mathcal{C}$ and soft constraints $\mathcal{D}$, the respective penalties are defined as:
\[
P_{CC} = \sum_{c \in \mathcal{C}} w_c g_c(S), \quad
P_{SC} = \sum_{c \in \mathcal{D}} w_c g_c(S)
\]

The final NCV score is given by:
\[
NCV = HSC \cdot \left(1 - (\alpha P_{CC} + \beta P_{SC})\right)
\]

where \( \alpha, \beta \in [0,1] \) control the influence of contextual and soft violations.

By construction, \( NCV \in [0,1] \). Any violation of a hard constraint forces \( NCV = 0 \), while contextual and soft constraints reduce the score proportionally.

NCV therefore complements SSS and OCS by ensuring that, in addition to semantic stability and structural consistency, the advisory remains compliant with critical operational constraints.

\subsection{Decision Readiness Level (DRL)}
\label{DRL}

The final stage of our framework is the \textbf{Decision Readiness Level (DRL)}. We chose a Fuzzy Inference System (Mamdani) for this layer because human trust is rarely binary. Instead of a rigid average, the DRL mimics expert reasoning: if a flight plan violates a safety rule (low NCV), it doesn't matter how 'stable' or 'consistent' it is—the fuzzy logic immediately demotes it to DRL-1 (Unsuitable). This approach enables ATAL to provide a calibrated, interpretable signal that indicates not only whether a model is confident, but also why it should or should not be trusted.

Rather than relying on a fixed linear aggregation, ATAL models decision readiness as a latent construct inferred from these indicators. To achieve this, the DRL is computed using a \textbf{fuzzy inference system (Mamdani)}, in which SSS, OCS, and NCV are treated as linguistic variables (e.g., low, medium, high).

The fuzzy inference process applies a set of expert-defined rules that encode operational reasoning, such as:

\begin{itemize}
    \item IF (SSS is high) AND (OCS is high) AND (NCV is high) THEN DRL is Decision-ready
    \item IF (NCV is low) THEN DRL is Unusable
    \item IF (SSS is medium) AND (OCS is medium) THEN DRL is Supervised
\end{itemize}

This rule-based formulation enables a gradual and interpretable mapping between reliability signals and decision authority, avoiding abrupt threshold effects typically associated with linear aggregation.

To preserve interpretability and provide a baseline comparison, a linear Trust Evidence Score (TES) can still be computed as:
\[
TES = w_s \, SSS + w_o \, OCS + w_n \, NCV,
\]
subject to $w_s + w_o + w_n = 1$, where weights are non-negative and domain-dependent. However, TES is not used directly for decision-making; instead, it serves as an auxiliary reference.

To account for potential misalignment between model-declared confidence and observed reliability, a calibration adjustment can be applied. Let $u_C \in [0,1]$ denote the declared confidence, and define the empirical confidence as:
\[
E = w_s \, SSS + w_o \, OCS + w_n \, NCV.
\]
The Confidence Calibration Index (CCI) is defined as:
\[
CCI = 1 - |u_C - E|.
\]

Finally, the DRL is derived through the aggregation of constituent scores.
To evaluate alternative decision-readiness formulations, we compared three distinct aggregation methods: a linear baseline score, and two fuzzy logic implementations. The latter utilize fuzzy aggregation followed by defuzzification (e.g., centroid method) to generate a continuous index in the range $[0,1]$ (see Section~\ref{sec:Experimental Setup}). This aggregated score is subsequently mapped onto discrete operational categories:

\begin{itemize}
    \item DRL-1: Unsuitable for use
    \item DRL-2: Informational / reference only
    \item DRL-3: Conditional execution under human supervision
    \item DRL-4: Fully actionable / decision-ready
\end{itemize}

This approach enables a more realistic representation of uncertainty and partial reliability, supporting calibrated human-in-the-loop or autonomous decision-making in safety-critical environments.

\begin{figure}[htbp]
    \centering
    \includegraphics[width=\columnwidth]{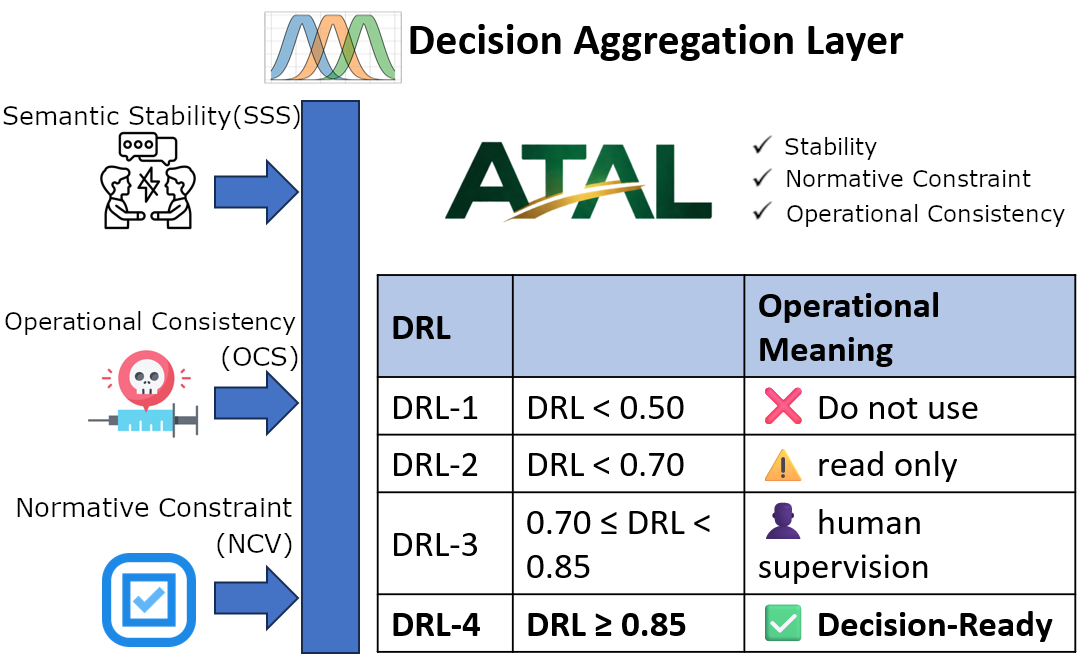}
    \caption{Decision Readiness Level (DRL) with fuzzy inference layer.}
    \label{fig:drl_pipeline}
\end{figure}

Figure \ref{fig:drl_pipeline} illustrates how ATAL integrates SSS, OCS, and NCV into the Decision Readiness Layer (DRL), where reliability signals are processed through a fuzzy inference layer and mapped into discrete operational levels.

\section{Experimental Results and Analysis}
\label{sec:Experimental Results and Analysis}

To evaluate the ATAL framework's efficacy in a realistic Air Traffic Management (ATM) environment, we developed twelve experimental scenarios (S01 to S12). These cases were specifically engineered to test the system’s sensitivity across varying levels of data integrity, safety, compliance, and decision-making implications.

The scenarios are categorized into three primary functional groups. First, the baseline and robustness scenarios (S01, S07, S08) represent ideal, complete, and consistent flight plans under prompt variations, serving as a positive control for high decision readiness (DRL-4). Second, input failure and structural integrity scenarios (S04, S05, S12) simulate critical errors—such as missing mandatory fields or invalid formatting—where the system must ensure immediate rejection of the output. Finally, ambiguity and operational context scenarios (S02, S03, S09, S10) explore nuanced failures, including the omission of required alternate aerodromes or the disregard of airspace restrictions, necessitating deeper normative analysis.

The experiments confirm that NCV behaves as a "safety gatekeeper."
Scenario S11 (Constraint Violation) exemplifies an "unsafe but consistent" output, where the AI generates a linguistically fluent and stable flight plan that nonetheless violates minimum separation standards or restricted zones. In such instances, the NCV component should act as the final safety filter, demoting the output to DRL-1 (Unsuitable) regardless of the AI’s perceived confidence or stability. Ultimately, these scenarios reflect real-world ATM challenges—such as weather dependency and strict adherence to ICAO Doc 4444, allowing the ATAL framework to provide clear, calibrated guidance to human operators by distinguishing between actionable autonomous suggestions and those requiring strict supervision or total rejection.

Table \ref{tab:scenario_atal} summarizes the scenarios, their technical characteristics, and their operational relevance, for instance, in S02 scenario, the AI’s failure to include an alternate aerodrome when required by destination weather directly compromises fuel reserves and mission safety.

\begin{table*}[t]
\centering
\caption{Operational Scenario Design and Relevance}
\label{tab:scenario_atal}
\begin{tabular}{|c|l|p{5cm}|p{5cm}|}
\hline
\textbf{Scenario} & \textbf{Type} & \textbf{Description} & \textbf{Operational Significance} \\ \hline

S01 & Valid baseline 
& Complete and consistent flight plan advisory generated by the model 
& Represents ideal conditions where the advisory can be directly executed without additional validation \\

S02 & Alternate required 
& Advisory omits or inconsistently handles alternate aerodrome requirements 
& Reflects real-world contingency planning gaps that may compromise safety under degraded conditions \\

S03 & Context-sensitive 
& Advisory must respect contextual constraints (e.g., airspace or restrictions) 
& Captures dependence on external operational conditions not explicitly encoded in the prompt \\

S04 & Incomplete input 
& Critical fields (e.g., departure, destination) are missing 
& Represents unusable outputs that must be rejected before any operational consideration \\

S05 & Conflicting input 
& Advisory contains contradictory information (e.g., inconsistent route or parameters) 
& Models ambiguity or internal inconsistency that can lead to unsafe execution \\

S06 & Unsafe but consistent 
& Structurally valid plan that violates safety constraints (e.g., restricted route) 
& Demonstrates risk of high-confidence but unsafe model outputs \\

S07 & Stable variation 
& Multiple outputs remain consistent under input perturbations 
& Reflects robustness of the model under minor variations \\

S08 & High consistency 
& Strong agreement across generated outputs 
& Indicates reliable advisory generation behavior \\

S09 & Ambiguous but resolvable 
& Multiple valid interpretations of the advisory exist 
& Represents decision-making under uncertainty requiring human judgment \\

S10 & Redundant but valid 
& Over-specified but internally consistent advisory 
& Reflects verbose but operationally acceptable outputs \\

S11 & Constraint violation 
& Advisory violates regulatory or operational constraint 
& Models realistic failure modes where compliance is critical \\

S12 & Structurally invalid 
& Output fails basic formatting or completeness rules 
& Represents outright invalid outputs that must be discarded \\

\hline
\end{tabular}
\end{table*}

\subsection{Experimental Setup}
\label{sec:Experimental Setup}

For each scenario, multiple advisory outputs were generated and processed through the ATAL pipeline. The Semantic Stability Score (SSS), Operational Consistency Score (OCS), and Normative Constraint Validation (NCV) were computed for each set of outputs.

The Decision Readiness Level (DRL) was then derived using three alternative formulations:

\begin{itemize}
    \item \textbf{Linear DRL (Baseline):} A weighted aggregation of SSS, OCS, and NCV, representing a traditional score-based approach.
    \item \textbf{Heuristic Fuzzy DRL:} A rule-based fuzzy approximation introducing gradual penalties for uncertainty.
    \item \textbf{Mamdani Fuzzy DRL:} A formal fuzzy inference system using linguistic variables and rule-based reasoning.
\end{itemize}

This setup enables a direct comparison between aggregation-based and inference-based decision strategies.

\subsection{Metric Behavior Analysis}

Before analyzing the aggregated Decision Readiness Level (DRL), it is useful to examine how each underlying metric behaves across the designed scenarios. Although SSS, OCS, and NCV are combined in the final decision layer, each captures a different dimension of advisory reliability.

Across the evaluated scenarios, the \textbf{Semantic Stability Score (SSS)} behaves largely as expected. It remains high in cases where outputs are robust under variation (e.g., S01, S07, S08), and decreases in scenarios involving ambiguity or structural variability (e.g., S03, S12). This pattern suggests that SSS is effectively capturing the sensitivity of model outputs to perturbations.

The \textbf{Operational Consistency Score (OCS)} exhibits a different behavior. It tends to remain high in structurally coherent scenarios, even when those scenarios are operationally unsafe (e.g., S06 and S11). This distinction is important: OCS measures internal agreement rather than correctness. In practice, this means that a model can be consistently wrong, and OCS alone will not penalize such cases. This reinforces the need for complementary validation mechanisms.

The \textbf{Normative Constraint Validation (NCV)} plays a more decisive role. In all scenarios involving missing information, conflicting inputs, or regulatory violations (e.g., S04, S05, and S12), NCV drops to zero, effectively invalidating the advisory regardless of its stability or consistency. This confirms that NCV correctly enforces structural and safety constraints as a gating mechanism.

Scenarios such as S11 highlight the interaction between these metrics. While both SSS and OCS remain high, indicating stable and internally consistent outputs, NCV penalizes the advisory due to constraint violations (i.e., non-compliance). 

Overall, these results indicate that no single metric is sufficient to characterize advisory reliability in isolation. Instead, SSS, OCS, and NCV provide complementary signals, and their combined interpretation is necessary to capture the range of behaviors observed in safety-critical contexts.

Table~\ref{tab:metric_summary} summarizes these patterns across all scenarios. While not exhaustive, it highlights the dominant behavior of each metric and reinforces its distinct roles.

In particular, the table shows that NCV is the primary determinant in invalid scenarios, whereas SSS and OCS may remain high even when the advisory is unsafe. This further motivates the need for an integrated decision layer to reconcile these signals.

\begin{table}[htbp]
\centering
\caption{Summary of Metric Behavior Across Scenarios}
\label{tab:metric_summary}
\begin{tabular}{|c|c|c|c|}
\hline
Metric & High Cases & Medium Cases & Low Cases \\ \hline
SSS & S01, S07, S08 & S03 & S12 \\
OCS & S01, S06, S11 & S09 & S05 \\
NCV & S01, S07 & S02, S09 & S04, S05, S12 \\
\hline
\end{tabular}
\end{table}

\subsection{DRL Results Overview}

\begin{figure*}[t]
    \centering
    \includegraphics[width=.95\textwidth]{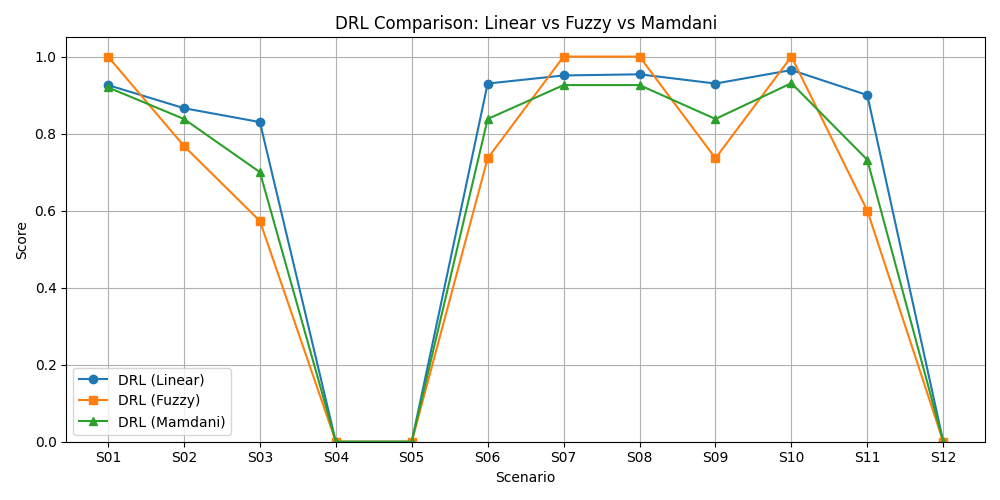}
    \caption{Comparison of DRL formulations: Linear (baseline), heuristic fuzzy, and Mamdani fuzzy inference.}
    \label{fig:drl_comparison_full}
\end{figure*}

Figure~\ref{fig:drl_comparison_full} presents the DRL scores obtained across all scenarios for the three evaluated models.

All models consistently assign a DRL score of zero to scenarios violating hard constraints (e.g., missing critical fields or structural inconsistencies), demonstrating correct enforcement of safety requirements.

\subsection{Comparative Analysis}

While agreement is observed in clearly valid and invalid cases, significant differences emerge in intermediate scenarios.

The \textbf{linear baseline} tends to produce higher DRL scores, indicating a more optimistic assessment of decision readiness. This behavior suggests limited sensitivity to partial inconsistencies or structural weaknesses.

In contrast, the \textbf{heuristic fuzzy model} applies stronger penalties in such cases, often resulting in lower DRL scores. While this improves robustness, it may lead to overly conservative decisions.

The \textbf{Mamdani fuzzy model} provides a balanced alternative. By leveraging linguistic rules and gradual inference, it produces smoother transitions between decision levels. This results in more realistic assessments in scenarios involving ambiguity, partial compliance, or conflicting inputs.

\subsection{Key Observations}

The experimental results highlight three important findings:

\begin{itemize}
    \item \textbf{Constraint enforcement:} NCV effectively dominates all models, ensuring that structurally invalid outputs are consistently rejected.

    \item \textbf{Sensitivity to uncertainty:} Linear aggregation lacks the capacity to account for uncertainty, frequently resulting in biased overestimations. Conversely, fuzzy approaches effectively incorporate uncertainty and degrees of reliability into the decision-making.
    
    \item \textbf{Inference advantage:} The Mamdani model achieves a desirable balance between robustness and sensitivity, avoiding both overconfidence and excessive conservatism.
\end{itemize}

\subsection{Discussion}

The comparative results reveal consistent patterns across the evaluated models, particularly in intermediate scenarios (e.g., S02, S03, and S09) where decision readiness is less clear-cut.
Partial inconsistencies or reduced structural validity did not significantly impact the final score.

A particularly revealing case is S11, which exhibits high consistency but reduced validity and is treated very differently by the models. While the linear formulation maintains high readiness scores, both fuzzy approaches reduce the DRL (with Mamdani being more balanced), reflecting the impact of constraint violations. 

Figure \ref{fig:drl_comparison_full} shows that the largest divergence between models occurs in intermediate scenarios such as S02, S03, S09, and S11, precisely where decision readiness is least obvious.

Overall, these observations support the view that decision readiness cannot be fully captured through simple aggregation. Instead, treating it as an inference problem enables a more realistic representation of uncertainty, particularly in safety-critical contexts.

\subsection{Limitations}

This evaluation is conducted using controlled and partially simulated scenarios, which limits the ability to fully capture the variability and unpredictability of real-world LLM outputs. Future work will focus on validating the ATAL framework using real operational data and live model interactions.

\section{Conclusion}
\label{sec:Conclusion}

This paper addressed a fundamental gap in the use of generative AI for safety-critical decision support: the absence of a structured mechanism to determine when AI-generated outputs can be trusted for operational use.

We introduced the AI Trust and Assurance Layer (ATAL) as a model-agnostic framework designed to bridge the gap between probabilistic AI outputs and deterministic operational requirements. By combining Semantic Stability (SSS), Operational Consistency (OCS), and Normative Constraint Validation (NCV), ATAL provides a multidimensional view of reliability that goes beyond traditional evaluation approaches focused on surface-level correctness or static benchmarks.

The experimental results highlight a key insight: reliability is not a single property. 
Models may exhibit apparent stability/consistency despite being operationally invalid.
This distinction, observed in scenarios such as S11, underscores the need to separate stability, consistency, and validity when assessing AI-assisted decisions.

The proposed Decision Readiness Level (DRL) addresses this challenge by translating these heterogeneous signals into an operationally meaningful decision layer.
Among the tested scenarios, the Mamdani fuzzy inference system generally provided a balanced trade-off between sensitivity and precision.

While the evaluation is based on controlled and partially simulated scenarios, it provides a structured and transparent validation of the proposed framework. Future work will focus on extending this validation to real operational data and on expanding the framework to incorporate adversarial robustness and longitudinal reliability analysis.

Ultimately, ATAL is not intended to replace human oversight, but to formalize and strengthen it. By transforming qualitative trust into measurable assurance signals, the framework supports more informed and calibrated decision-making in environments where errors are not merely undesirable but unacceptable.

% Using the content of a pre-compiled .bbl file below...


\begin{thebibliography}{10}
\providecommand{\url}[1]{#1}
\csname url@samestyle\endcsname
\providecommand{\newblock}{\relax}
\providecommand{\bibinfo}[2]{#2}
\providecommand{\BIBentrySTDinterwordspacing}{\spaceskip=0pt\relax}
\providecommand{\BIBentryALTinterwordstretchfactor}{4}
\providecommand{\BIBentryALTinterwordspacing}{\spaceskip=\fontdimen2\font plus
\BIBentryALTinterwordstretchfactor\fontdimen3\font minus \fontdimen4\font\relax}
\providecommand{\BIBforeignlanguage}[2]{{%
\expandafter\ifx\csname l@#1\endcsname\relax
\typeout{** WARNING: IEEEtran.bst: No hyphenation pattern has been}%
\typeout{** loaded for the language `#1'. Using the pattern for}%
\typeout{** the default language instead.}%
\else
\language=\csname l@#1\endcsname
\fi
#2}}
\providecommand{\BIBdecl}{\relax}
\BIBdecl

\bibitem{qin2025}
Y.~Qin \emph{et~al.}, ``\BIBforeignlanguage{en}{Tool learning with foundation models},'' \emph{\BIBforeignlanguage{en}{ACM Comput. Surveys}}, vol.~57, no.~4, pp. 1--40, Apr. 2025, doi: 10.1145/3704435.

\bibitem{yao2023}
S.~Yao \emph{et~al.}, ``{ReAct}: Synergizing reasoning and acting in language models,'' in \emph{Proc. 11th Int. Conf. Learn. Represent.}, Kigali, Rwanda, May 2023.

\bibitem{park2023}
J.~S. Park, J.~O'Brien, C.~J. Cai, M.~R. Morris, P.~Liang, and M.~S. Bernstein, ``\BIBforeignlanguage{en}{Generative agents: Interactive simulacra of human behavior},'' in \emph{\BIBforeignlanguage{en}{Proc. 36th Annu. ACM Symp. User Interface Softw. Technol.}}\hskip 1em plus 0.5em minus 0.4em\relax San Francisco, CA, USA: ACM, Oct. 2023, pp. 1--22, doi: 10.1145/3586183.3606763. ISBN 979-8-4007-0132-0.

\bibitem{brown2020}
T.~B. Brown \emph{et~al.}, ``Language models are few-shot learners,'' in \emph{Proc. 34th Int. Conf. Neural Inf. Process. Syst.}, ser. Advances in Neural Information Processing Systems 33.\hskip 1em plus 0.5em minus 0.4em\relax Red Hook, NY, USA: Curran Assoc., Inc., 2020, pp. 1877--1901. ISBN 9781713829546.

\bibitem{rafael2020}
C.~Raffel \emph{et~al.}, ``Exploring the limits of transfer learning with a unified text-to-text transformer,'' \emph{J. Mach. Learn. Res.}, vol.~21, no. 140, pp. 5485--5551, Jan. 2020.

\bibitem{zou2023}
A.~Zou, Z.~Wang, N.~Carlini, M.~Nasr, J.~Z. Kolter, and M.~Fredrikson, ``Universal and transferable adversarial attacks on aligned language models,'' \emph{arXiv}, arXiv:2307.15043 [cs.CL], Dec. 2023.

\bibitem{liu2025}
Y.~Liu \emph{et~al.}, ``Prompt injection attack against {LLM}-integrated applications,'' \emph{arXiv}, arXiv:2306.05499 [cs.CR], Dec. 2025.

\bibitem{Damacena2026}
J.~Damacena~Duarte \emph{et~al.}, ``A systematic review of prompt injection attacks on large language models: Trends, taxonomy, evaluation, defenses, and opportunities,'' \emph{IEEE Access}, vol.~14, pp. 12\,875--12\,899, 2026, doi: 10.1109/ACCESS.2026.3656849.

\bibitem{brown1992}
P.~F. Brown, V.~J. Della~Pietra, P.~V. deSouza, J.~C. Lai, and R.~L. Mercer, ``Class-based n-gram models of natural language,'' \emph{Comput. Linguistics}, vol.~18, no.~4, pp. 467--479, Dec. 1992.

\bibitem{ji2023}
Z.~Ji \emph{et~al.}, ``\BIBforeignlanguage{en}{Survey of hallucination in natural language generation},'' \emph{\BIBforeignlanguage{en}{ACM Comput. Surveys}}, vol.~55, no.~12, pp. 1--38, Dec. 2023, doi: 10.1145/3571730.

\bibitem{srivastava2023}
A.~Srivastava \emph{et~al.}, ``Beyond the imitation game: Quantifying and extrapolating the capabilities of language models,'' \emph{Trans. Mach. Learn. Res.}, 2023.

\bibitem{liang2023}
P.~Liang \emph{et~al.}, ``Holistic evaluation of language models,'' \emph{Trans. Mach. Learn. Res.}, 2023.

\bibitem{lin2022}
S.~Lin, J.~Hilton, and O.~Evans, ``\BIBforeignlanguage{en}{{TruthfulQA}: Measuring how models mimic human falsehoods},'' in \emph{\BIBforeignlanguage{en}{Proc. 60th Annu. Meeting Assoc. Comput. Linguistics}}, vol.~1.\hskip 1em plus 0.5em minus 0.4em\relax Dublin, Ireland: Assoc. Comput. Linguistics, May 2022, pp. 3214--3252, doi: 10.18653/v1/2022.acl-long.229.

\bibitem{perez2022}
E.~Perez \emph{et~al.}, ``Red teaming language models with language models,'' in \emph{Proc. 2022 Conf. Empirical Methods Natural Lang. Process.}, Y.~Goldberg, Z.~Kozareva, and Y.~Zhang, Eds.\hskip 1em plus 0.5em minus 0.4em\relax Abu Dhabi, United Arab Emirates: Assoc. Comput. Linguistics, Dec. 2022, pp. 3419--3448, doi: 10.18653/v1/2022.emnlp-main.225.

\bibitem{ICAO44442016}
\BIBentryALTinterwordspacing
{International Civil Aviation Organization}, ``{Procedures for Air Navigation Services — Air Traffic Management (PANS-ATM)},'' Int. Civil Aviation Org., Montréal, QC, Canada, Manual Doc 4444, 2016, incorporating Amendments 1--7. [Online]. Available: \url{https://store.icao.int/en/procedures-for-air-navigation-services-air-traffic-management-doc-4444}.
\BIBentrySTDinterwordspacing

\bibitem{rtca2011}
{RTCA} and {EUROCAE}, ``Software considerations in airborne systems and equipment certification,'' {RTCA Inc.} and {European Organisation for Civil Aviation Equipment}, Washington, DC, Standard {DO-178C} / {ED-12C}, 2011.

\bibitem{nistai2023}
E.~Tabassi, ``{Artificial Intelligence Risk Management Framework (AI RMF 1.0)},'' Nat. Inst. Standards Technol., Gaithersburg, MD, Report {NIST AI 100-1}, Jan. 2023, doi: 10.6028/NIST.AI.100-1.

\bibitem{euai2024}
\BIBentryALTinterwordspacing
{European Parliament and Council of the European Union}, ``Regulation ({EU}) 2024/1689 of the {European Parliament} and of the {Council} of 13 june 2024 laying down harmonised rules on artificial intelligence and amending regulations ({EC}) no 300/2008, ({EU}) no 167/2013, ({EU}) no 168/2013, ({EU}) 2018/858, ({EU}) 2018/1139 and ({EU}) 2019/2144 and directives 2014/90/{EU}, ({EU}) 2016/797 and ({EU}) 2020/1828 ({Artificial Intelligence Act}),'' {Official Journal of the European Union, L 2024/1689}, July 2024. [Online]. Available: \url{https://eur-lex.europa.eu/eli/reg/2024/1689/oj}.
\BIBentrySTDinterwordspacing

\bibitem{iec2010}
{IEC}, ``Functional safety of electrical/electronic/programmable electronic safety-related systems,'' Int. Electrotech. Comm., Geneva, Switzerland, Standard IEC 61508, 2010.

\bibitem{christiano2017}
P.~F. Christiano, J.~Leike, T.~B. Brown, M.~Martic, S.~Legg, and D.~Amodei, ``Deep reinforcement learning from human preferences,'' in \emph{Proc. 31st Int. Conf. Neural Inf. Process. Syst.}, vol.~30.\hskip 1em plus 0.5em minus 0.4em\relax Red Hook, NY, USA: Curran Assoc., Inc., 2017, pp. 4302--4310. ISBN 9781510860964.

\bibitem{Ouyang2022}
L.~Ouyang \emph{et~al.}, ``Training language models to follow instructions with human feedback,'' in \emph{Proc. 36th Int. Conf. Neural Inf. Process. Syst.}, no. 2011.\hskip 1em plus 0.5em minus 0.4em\relax Red Hook, NY, USA: Curran Assoc., Inc., 2022, pp. 27\,730--27\,744. ISBN 9781713871088.

\bibitem{guo2017}
C.~Guo, G.~Pleiss, Y.~Sun, and K.~Q. Weinberger, ``On calibration of modern neural networks,'' in \emph{Proc. 34th Int. Conf. Mach. Learn.}, ser. ICML'17, vol.~70.\hskip 1em plus 0.5em minus 0.4em\relax JMLR.org, Aug. 2017, pp. 1321--1330.

\bibitem{hendrycks2017}
D.~Hendrycks and K.~Gimpel, ``A baseline for detecting misclassified and out-of-distribution examples in neural networks,'' in \emph{Int. Conf. Learn. Represent.}, 2017.

\bibitem{goodfellow2015}
I.~J. Goodfellow, J.~Shlens, and C.~Szegedy, ``Explaining and harnessing adversarial examples,'' in \emph{Proc. 3rd Int. Conf. Learn. Represent.}, Y.~Bengio and Y.~LeCun, Eds., San Diego, CA, USA, 2015.

\bibitem{arrieta2020}
A.~B. Arrieta \emph{et~al.}, ``Explainable {A}rtificial {I}ntelligence ({XAI}): Concepts, taxonomies, opportunities and challenges toward responsible {AI},'' \emph{Inf. Fusion}, vol.~58, pp. 82--115, 2020, doi: 10.1016/j.inffus.2019.12.012.

\bibitem{dwivedi2023}
R.~Dwivedi \emph{et~al.}, ``\BIBforeignlanguage{en}{Explainable {AI} ({XAI}): Core ideas, techniques, and solutions},'' \emph{\BIBforeignlanguage{en}{ACM Comput. Surveys}}, vol.~55, no.~9, pp. 1--33, Sep. 2023, doi: 10.1145/3561048.

\bibitem{gunning2021}
D.~Gunning, E.~Vorm, J.~Y. Wang, and M.~Turek, ``\BIBforeignlanguage{en}{{DARPA}'s explainable {AI} ({XAI}) program: A retrospective},'' \emph{\BIBforeignlanguage{en}{Appl. AI Lett.}}, vol.~2, no.~4, Dec. 2021, doi: 10.1002/ail2.61.

\end{thebibliography}
\end{document}